\documentclass[10pt,twocolumn,letterpaper]{article}

\usepackage{arxiv}
\usepackage{times}
\usepackage{epsfig}
\usepackage{graphicx}
\usepackage{amsmath}
\usepackage{amssymb}
\usepackage{tabularx}
\usepackage{booktabs}
\usepackage{enumerate}
\usepackage{lipsum}
\usepackage{float}
\usepackage{subfigure}
\usepackage{array}
\usepackage{breqn}
\usepackage{float}
\usepackage{algorithm}
\usepackage{algpseudocode}
\usepackage{capt-of}
\usepackage{pdfpages}
\usepackage{enumerate}
\usepackage{footnote}
\usepackage{multirow}
\usepackage{makecell}
\usepackage{textcomp}
\usepackage[usenames,dvipsnames]{color}
\usepackage{colortbl}
\definecolor{mygray}{gray}{.9}

\usepackage[pagebackref=true,breaklinks=true,letterpaper=true,colorlinks,bookmarks=false]{hyperref}

\begin{document}

\makeatletter
\renewcommand{\@fnsymbol}[1]{\ensuremath{%
   \ifcase#1\or \# \or *\or {*}*\or
   \mathsection\or \mathparagraph\or \|\or \star\or
   \star\star\or {\star\star}\star \else\@ctrerr\fi}}
\makeatother

\title{Attribute Guided Unpaired Image-to-Image Translation\\with Semi-supervised Learning}

\author{
    Xinyang Li$^{1}$\thanks{Contributed Equally.} \ ,
    Jie Hu$^{1\#}$,
    Shengchuan Zhang$^1$,
    Xiaopeng Hong$^2$, \\
    Qixiang Ye$^3$,
    Chenglin Wu$^4$,
    and Rongrong Ji$^{1}$\thanks{Corresponding Author.}\\
    $^1$Xiamen University,
    $^2$Xi'an Jiaotong University,
    $^3$University of Chinese Academy of Sciences,
    $^4$Fuzhi.ai\\
    {\tt\small \{Imlixinyang,hujie.cpp,ether.wcl\}@gmail.com, \{rrji, zsc\_2016\}@xmu.edu.cn}\\
    {\tt\small hongxiaopeng@mail.xjtu.edu.cn, qxye@ucas.ac.cn}
}
\maketitle

\begin{abstract}
Unpaired Image-to-Image Translation (UIT) focuses on translating images among different domains by using unpaired data, which has received increasing research focus due to its practical usage.
However, existing UIT schemes defect in the need of supervised training, as well as the lack of encoding domain information.
In this paper, we propose an Attribute Guided UIT model termed AGUIT to tackle these two challenges.
AGUIT considers multi-modal and multi-domain tasks of UIT jointly with a novel semi-supervised setting, which also merits in representation disentanglement and fine control of outputs.
Especially, AGUIT benefits from two-fold:
(1) It adopts a novel semi-supervised learning process by translating attributes of labeled data to unlabeled data, and then reconstructing the unlabeled data by a cycle consistency operation.
(2) It decomposes image representation into domain-invariant content code and domain-specific style code.
The redesigned style code embeds image style into two variables drawn from standard Gaussian distribution and the distribution of domain label, which facilitates the fine control of translation due to the continuity of both variables.
Finally, we introduce a new challenge, \ie, disentangled transfer, for UIT models, which adopts the disentangled representation to translate data less related with the training set.
Extensive experiments demonstrate the capacity of AGUIT over existing state-of-the-art models.
\end{abstract}


\section{Introduction}

Image-to-image translation aims to learn the mapping between images among different domains, which has drawn increasing research attention.
Many computer vision tasks can be modeled as an image-to-image translation problem, such as colorization \cite{zhang2016colorful}, super resolution \cite{ledig2017photo,wu2017srpgan}, semantic synthesis \cite{chen2017photographic} and domain adaption \cite{hoffman2017cycada}.
\begin{figure}[!t]
\centering
\includegraphics[width=3.2in]{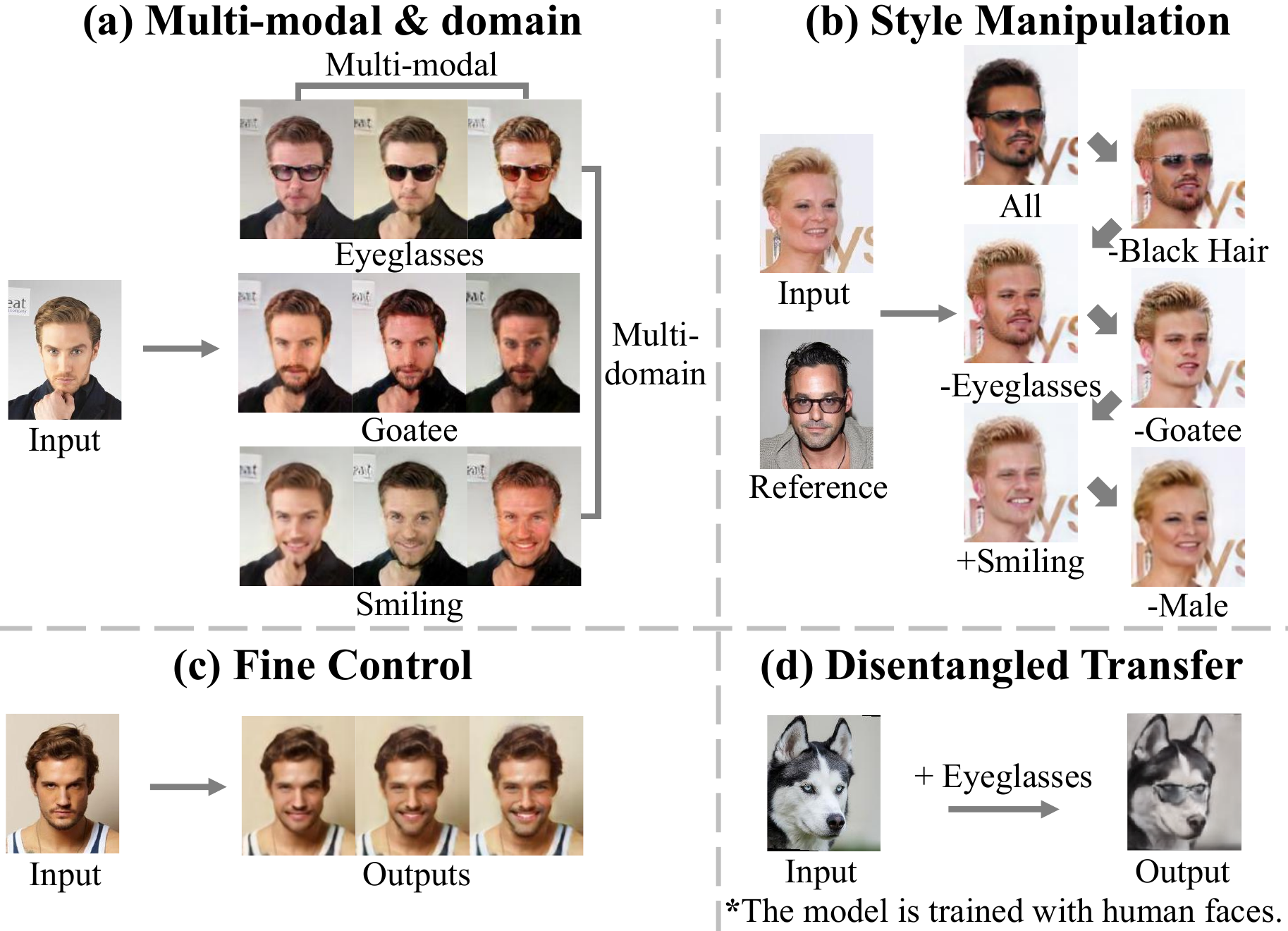}
\caption{
Examples of UIT tasks accomplished by AGUIT.
(a) Multi-modal and multi-domain translation are achieved jointly.
(b) Style code of reference image can be manipulated while retaining the content of input image.
(c) The outputs can be fine controlled.
(d) An example of disentangled transfer.
}
\label{fig11}
\end{figure}
Earliest works in image-to-image translation can be referred to translate paired images (\ie, Pix2Pix \cite{isola2017image}), in which every image in the source domain should have the corresponding paired image in the target domain for training.
However, it is expensive to collect paired images in practice.
To overcome this limitation, several methods were proposed for unpaired image-to-image translation (UIT), such as CycleGAN \cite{zhu2017unpaired}, DualGAN \cite{yi2017dualgan} and UNIT \cite{liu2017unsupervised}.
\begin{table*}
\begin{center}
\scalebox{0.72}[0.72]{
\begin{tabular}{ccccccccc}
\hline 
Merits & Unpaired & Multi-modal & Multi-domain & Single Model & Fine Control &  Disentanglement & Semi Supervised &  Disentangled Transfer \\ \hline
Pix2Pix \cite{isola2017image} & - & - & - & - & - & - & - & -\\\rowcolor{mygray}
BicycleGAN \cite{zhu2017toward} & - & ${\surd}$ & - & - & - & - & - & - \\
CycleGAN \cite{zhu2017unpaired} & ${\surd}$ & - & - & - & - & - & - & - \\\rowcolor{mygray}
MUNIT \cite{huang2018multimodal} / DRIT \cite{lee2018diverse} & ${\surd}$ & ${\surd}$ & - & - & - & - & - & - \\
StarGAN \cite{Choi2017StarGAN} & ${\surd}$ &  - & ${\surd}$ & ${\surd}$ & - & - & - & -  \\\rowcolor{mygray}
GANimation \cite{pumarola2018ganimation} & ${\surd}$ & - & ${\surd}$ & ${\surd}$ & ${\surd}$ & - & - & - \\
SMIT \cite{romero2018smit} & ${\surd}$ &  ${\surd}$ & ${\surd}$ & ${\surd}$ & - & -  & - & -  \\\rowcolor{mygray}
\textbf{AGUIT} & ${\surd}$ & ${\surd}$ & ${\surd}$ & ${\surd}$ & ${\surd}$ & ${\surd}$ & ${\surd}$ & ${\surd}$ \\ \hline
\end{tabular}}
\end{center}
\vspace{-5px}
\caption{
The merits of AGUIT compared with the existing state-of-the-art image-to-image translation models.
}
\vspace{-10px}
\label{tab1}
\end{table*}

UIT involves two basic tasks, \ie, multi-domain translation and multi-modal translation.
The former aims to use a single model to achieve translations across multiple domains.
The latter aims to generate diversified outputs while retaining the content information.
For multi-domain translation, StarGAN \cite{Choi2017StarGAN} was proposed to input the labels together with images into the model to guide the translation.
However, due to the fixed and discrete labels, it cannot generate diverse outputs.
GANimation \cite{pumarola2018ganimation} describes the labels in a continuous manifold, which however severely relies on the Action Units annotations.
For multi-modal translation, MUNIT \cite{huang2018multimodal} and DRIT \cite{lee2018diverse} were proposed, in which the style code is drawn from the Gaussian distribution to enrich the outputs.
However, these two methods need different models for different translations.
Recently, SMIT \cite{romero2018smit} is proposed to achieve these two tasks jointly, in which the images with labels and additional style noises are input to the generator.
However, it cannot control the result finely since the domain information is not encoded.
Overall, these state-of-the-art UIT models are defect in the need of supervised training, as well as a suitable and explicit encoding for domain information.
We argue that an ideal UIT model should have the following merits:
First, the unlabeled data should be incorporated into training process to achieve \emph{semi-supervised learning}, so as to reduce the requirement of expensive label annotations.
Second, the domain information (or attributes) of images should be explicitly learned to \emph{finely control} the result, so as to enhance the translation flexibility.
Third, the learned representation should be \emph{disentangled}, so the inputs can be translated by specific semantic information to improve the translation interpretability.
In this paper, we achieve the above goals in a unified framework by proposing an Attribute Guided Unpaired Image-to-image Translation (AGUIT) model.
Two innovative designs are presented:
(1) It adopts a novel semi-supervised learning process by translating attributes of labeled data to unlabeled data, and then reconstructing the unlabeled data with a cycle consistency operation.
(2) It decomposes image representation into domain-invariant content code and domain-specific style code.
The style code embeds image style into two variables drawn from standard Gaussian distribution and domain label distribution, which enhances the controllability of translation due to the continuity of both distributions.
Particularly, the standard Gaussian distribution encourages unsupervised disentanglement, since its every dimension is independent with each other.
And the domain label distribution is disentangled naturally and forced to be continuous by our training scheme.
Finally, we introduce a \emph{disentangled transfer}, which is a new challenge for UIT to adopt the disentangled representation to translate data less related with training set.
The examples of UIT tasks accomplished by AGUIT are shown in Fig.~\ref{fig11}.
Extensive experiments demonstrate the capacity of AGUIT over the state-of-the-art image-to-image translation models.
First, AGUIT performs well on the basic tasks (\ie, multi-domain translation and multi-modal translation) of UIT.
Second, qualitative and quantitative evaluations reveal the benefits of our semi-supervised scheme.
Third, AGUIT works well for high-level UIT tasks such as style manipulation and fine control.
Finally, AGUIT carries out disentangled transfer, which is a new task introduced to image-to-image translation.
The merits of AGUIT are summarized as Tab.~\ref{tab1}.
Our contributions are listed as below:
\begin{itemize}
\vspace{-5px}
\item The proposed AGUIT, which employs a novel semi-supervised learning process and a novel style code for translation.
To our best knowledge, AGUIT is the first model to consider multi-modal translation and multi-domain translation with a semi-supervised setting, and achieves disentanglement of representation as well as fine control of outputs for UIT.
\vspace{-5px}
\item We introduce a new translation task termed disentangled transfer, in which the disentangled representation is adopted to translate data less related with training set.
\vspace{-5px}
\item Extensive experiments demonstrate the capacity of AGUIT over the existing state-of-the-art image-to-image translation models.
\vspace{-5px}
\end{itemize}
The rest of this paper is organized as follows.
Sec.~\ref{sec:RW} reviews the related work.
The proposed AGUIT is introduced in Sec.~\ref{sec:AGUIT}.
Qualitative and quantitative experiments are given in Sec.~\ref{sec:Experiments}.
%
%
Finally, we conclude this work in Sec.~\ref{sec:Conclusion}.
%
\section{Related Work}
\label{sec:RW}
\textbf{Generative Adversarial Network.} GAN \cite{goodfellow2014generative} has achieved remarkable results.
In the training stage of GAN, the discriminator tries to distinguish the outputs of generator and the real distribution.
On the contrary, the generator tries to fool the discriminator.
After training, the generator can produce outputs which are similar to the real samples.
In our model, GAN is used to align the labels for inputs, constrain the content code to common space, and make the style code and content code be separable.
\begin{figure*}[!t]
\centering
\includegraphics[width=6.4in]{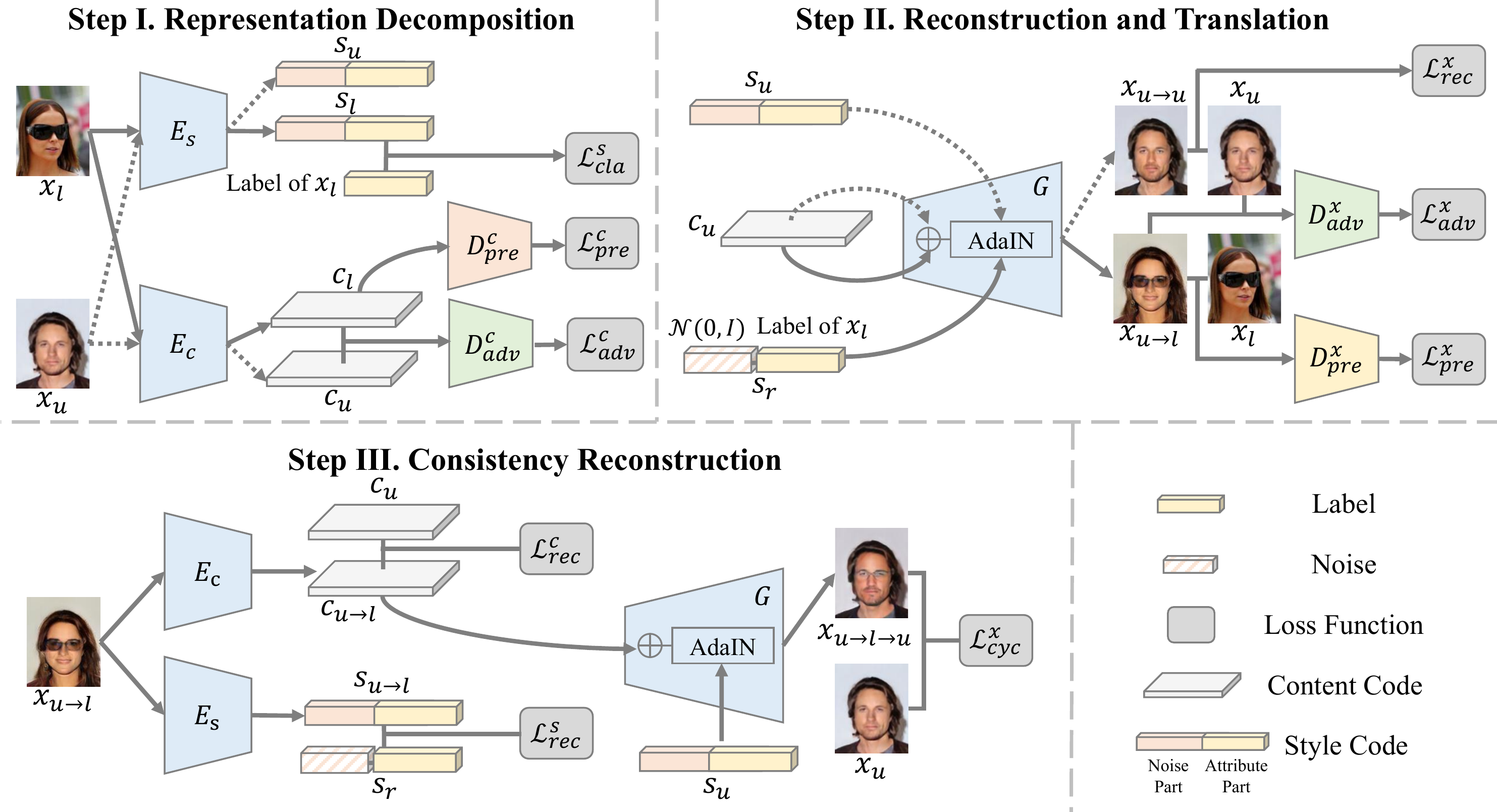}
\caption{
The flowchart of training AGUIT.
First, the representation of image $x_l$ and image $x_u$ is decomposed into style code $s_l, s_u$ and content code $c_l, c_u$ by style encoder $E_s$ and content encoder $E_c$.
The label $l=[1,-1,1,...]$ corresponds to the attributes [glasses, male, smile,...] and 1 means $x_l$ has the corresponding attribute, -1 is opposite.
Second, the generator $G$ decodes $s_u, c_u$ to reconstruct $x_u$, and decodes $s_r, c_u$ to translate attributes of $x_l$ to $x_u$.
The $s_r$ consists of noise $z\sim\mathcal{N}(0,I)$ and label $l$.
The reconstructed and translated images are denoted as $x_{u\to u}$ and $x_{u\to l}$, respectively.
Third, $x_u$ is cyclically reconstructed by using $c_{u\to l}$ and $s_u$.
}
\label{fig1}
\end{figure*}

\textbf{Semi-supervised GANs.} Several recent works leveraged GANs for semi-supervised learning of classification models.
The works of \cite{springenberg2015unsupervised, odena2017conditional} train a discriminator to classify input into different classes.
The work of \cite{chongxuan2017triple} introduces a separate discriminator and classifier models. 
Other approaches incorporate inference models to predict missing labeled features \cite{denton2017unsupervised} or harness the joint distribution of labels and data matching \cite{gan2017triangle}.
Recently, the setting of training a classifier from a few labels is introducing to generative model.
The work of \cite{lucic2019highfidelity} learns a generative model from a few labels by clustering deep features. 
Unlike the above works, we focus on exploiting semi-supervised learning for image-to-image translation task.
\textbf{Image-to-Image Translation.} Image-to-image translation tasks have attracted increasing focus.
For instance, Pix2pix \cite{isola2017image} achieves translation based on paired image.
To use unpaired images, CycleGAN \cite{zhu2017unpaired}, DiscoGAN \cite{kim2017learning}, DualGAN \cite{yi2017dualgan} and UNIT \cite{liu2017unsupervised} are proposed.
For multi-domain translation, ComboGAN \cite{anoosheh2018combogan} learns multiple generators and StarGAN \cite{Choi2017StarGAN} reduces them to a single one by inputting the target labels and images together.
GANimation \cite{pumarola2018ganimation} describes the labels in a continuous manifold.
CerfGAN \cite{liu2018cerf} adopts a multi-class discriminator to enable the generator to translate images with high domain shifts.
For multi-modal translation, BicycleGAN \cite{zhu2017toward} extends pix2pix by learning a stochastic mapping from source to target.
MUNIT \cite{huang2018multimodal} and DRIT \cite{lee2018diverse} decompose the image representation into style code and content code, and then decode back to translated images.
Augmented CycleGAN \cite{Almahairi2018Augmented} learns many-to-many translation by using the stochastic mappings.
Recently, SMIT \cite{romero2018smit} is proposed to solve these basic tasks jointly, in which the images with labels and additional style noises are input to the generator.
Unlike existing methods, AGUIT encodes the attributes into the style code to conduct high-level translation tasks.

\textbf{Representation Disentanglement.} Representation disentanglement aims at disentangling and explaining the learned representation.
The methods can be divided into two categories, \ie, supervised disentanglement and unsupervised disentanglement.
Supervised disentanglement methods \cite{cheung2014discovering,mathieu2016disentangling,makhzani2015adversarial,kingma2014semi,Liu2018Detach,liu2018unified} make use of labeled data while unsupervised disentanglement methods \cite{chen2016infogan,higgins2017beta,burgess2018understanding,kim2018disentangling,chen2018isolating} learn the properties from unlabeled data.
The proposed AGUIT can conduct both supervised and unsupervised disentanglement by the proposed ingenious style code.
%

\section{The Model of AGUIT}
\label{sec:AGUIT}
\subsection{Problem Formulation}
Let $(\mathcal{X}_l, L)$ denote the pairs of images with corresponding attribute labels and $\mathcal{X}_u$ denote the images without attribute labels.
The image representation is decomposed into the style code $\mathcal{S}$ and the content code $\mathcal{C}$.
Our goal is to train a model for UIT by using $(\mathcal{X}_l, L)$ and $\mathcal{X}_u$, in which $\mathcal{C}$ retains the content of objects and $\mathcal{S}$ learns disentanglement for attributes.
After training, by manipulating $\mathcal{S}$, the expected attributes should be translated to the outputs.
The flowchart of training the proposed AGUIT for tackling this problem is shown in Fig.~\ref{fig1}.
\subsection{Representation Decomposition}
As the step I in Fig.~\ref{fig1}, we decompose the image representation of $\mathcal{X}_l$ into $\mathcal{S}_l$ and $\mathcal{C}_l$ by style encoder $E_s$ and content encoder $E_c$.
The same operation is conducted on $\mathcal{X}_u$ to get $\mathcal{S}_u$ and $\mathcal{C}_u$.
Notably, $\mathcal{S}$ consists of the noise part and the attribute part.
\textbf{Style Classifying Loss.} To encourage continuity for the attribute part of style code, we enforce it to be close to $L$ in the Euclidean space.
This design enhances the capability of supervised disentanglement, because the continuous code can be used for fine control.
The style classifying loss is defined as:

\begin{equation}
\label{eq1}
\mathcal{L}_{cla}^s = \mathbb{E}_{(x_l, l) \in (\mathcal{X}_l, L)}[\Arrowvert E_s(x_l)_l - l \Arrowvert_2],
\end{equation}
where $E_s(x_l)_l$ denotes the attribute part of style code.
%

\textbf{Content Confusing Loss.} To constrain the content code to a common space, we utilize a discriminator $D_{adv}^c$ to train $E_c$.
In the training process, $D_{adv}^c$ tries to distinguish $\mathcal{C}_l$ and $\mathcal{C}_u$, while $E_c$ learns a common representation for them.
The content confusing loss is as follows:
\begin{equation}
\begin{split}
\mathcal{L}_{adv}^c = \mathbb{E}_{x_l\in \mathcal{X}_l,x_u\in\mathcal{X}_u}[&\log( D_{adv}^c(E_c(x_u)))\\+&\log(1- D_{adv}^c(E_c(x_l)))].
\end{split}
\end{equation}
\textbf{Content-Style Separating Loss.} To make the style code and content code independent to each other, we introduce a predictor $D_{pre}^c$ as inspired by PM \cite{schmidhuber1992learning}.
Because the style code is not stable in the early training phase, we let $D_{pre}^c$ directly predict the image labels conditioned on content code and prevent $E_c$ from learning the information of image labels.
Therefore, the content-style separating loss is defined as:
\begin{equation}
\mathcal{L}_{pre}^c = \mathbb{E}_{(x_l, l) \in (\mathcal{X}_l, L)}[\log(D_{pre}^c(l|E_c(x_l)))].
\end{equation}
\subsection{Reconstruction and Translation}
As the step II in Fig.~\ref{fig1}, we reconstruct $\mathcal{X}_u$ by inputting $\mathcal{S}_u$ and $\mathcal{C}_u$ to the generator $G$ with AdaIN \cite{huang2017arbitrary}.
Similarly, the translation of $\mathcal{X}_{u}$ is done by inputting $\mathcal{S}_r$ and $\mathcal{C}_u$ to $G$.
The $\mathcal{S}_r$ consists of the random noise $Z\sim\mathcal{N}(0, I)$ and the attribute label $L$ of $\mathcal{X}_l$.
We denote the reconstructed images as $\mathcal{X}_{u \to u}$ and the translated images as $\mathcal{X}_{u \to l}$.
\textbf{Image Reconstructing Loss.} To guarantee that the learned representation can be decoded to generate target images, we use an image reconstructing loss to train the translator (\ie, $E_s, E_c$ and $G$):
\begin{equation}
\mathcal{L}_{rec}^x =\mathbb{E}_{x_u\in\mathcal{X}_u}[\Arrowvert x_u - G(E_c(x_u),E_s(x_u))\Arrowvert _1].
\end{equation}
\textbf{Image Adversarial Loss.} To make the translator translate attributes $L$ to $\mathcal{X}_u$ while retaining the content, we use a discriminator $D_{adv}^x$ which attempts to distinguish $\mathcal{X}_u$ and $\mathcal{X}_{u \to l}$.
The translator tries to fool $D_{adv}^x$ by generating $\mathcal{X}_{u \to l}$ with the same content of $\mathcal{X}_u$.
Then, the image adversarial loss is defined as:
\begin{equation}
\begin{split}
\mathcal{L}_{adv}^x = \mathbb{E}_{x_u\in\mathcal{X}_u, x_l\in\mathcal{X}_l}[\log( D_{adv}^x&(G(E_c(x_u),s_r)))\\+\log&(1 -D_{adv}^x(x_l))],
\end{split}
\end{equation}
where $s_r\in\mathcal{S}_r$ contains a random noise $z\sim\mathcal{N}(0, I)$ and the attribute label $l\in L$ of $x_l$.
\textbf{Image Classifying Loss.} To make $\mathcal{X}_{u \to l}$ have the same attributes as $\mathcal{X}_l$, we apply a classifier $D_{pre}^x$ inspired by AC-GAN \cite{odena2017conditional}.
In the process, $D_{pre}^x$ attempts to predict $\mathcal{X}_l$ under the supervision of $L$, and the translator also tries to generate images satisfying the given label $L$.
Therefore, the domain classifying loss for $D_{pre}^x$ is defined as:
\begin{equation}
\mathcal{L}_{pre}^{x,D} = \mathbb{E}_{(x_l, l)\in(\mathcal{X}_l, L)}[\log(D_{pre}^{x}(l|x_l))].
\end{equation}
The loss for translator is defined as:
\begin{equation}
\mathcal{L}_{pre}^{x,G} = \mathbb{E}_{x_u\in\mathcal{X}_u}[\log(D_{pre}^{x}(l|G(E_c(x_u),s_r)))].
\end{equation}
\subsection{Consistency Reconstruction}
As the step III in Fig.~\ref{fig1}, we cyclically reconstruct the unlabeled images $\mathcal{X}_{u}$ with the style code $\mathcal{S}_{u \to l}$ and content code $\mathcal{C}_{u \to l}$ of $\mathcal{X}_{u \to l}$ encoded by $E_s$ and $E_c$.
\begin{algorithm}[t]
\caption{Process of Training AGUIT}
\label{alg1}
\hspace*{0.02in} {\bf Input:} %
The images with domain labels: $(\mathcal{X}_l, L)$. The images without requirement of labels: $\mathcal{X}_u$.
\\
\hspace*{0.02in} {\bf Output:} %
The learned style encoder $E_s$, content encoder $E_c$, and generator $G$.
\begin{algorithmic}[1]
\While{not convergence}
    \State Get $\mathcal{S}_l, \mathcal{C}_l$ by $E_s(\mathcal{X}_l), E_c(\mathcal{X}_l)$.
    \State Get $\mathcal{S}_u, \mathcal{C}_u$ by $E_s(\mathcal{X}_u), E_c(\mathcal{X}_u)$.
    \State Reconstruct $\mathcal{X}_u$: $\mathcal{X}_{u\to u} = G(\mathcal{C}_u, \mathcal{S}_u)$.
    \State Construct random style code: $\mathcal{S}_r = [\mathcal{N}(0, I), L]$.
    \State Translate $\mathcal{X}_u$ under style $\mathcal{S}_r$: $\mathcal{X}_{u \to l} = G(\mathcal{C}_u, \mathcal{S}_r)$.
    \State Get $\mathcal{S}_{u\to l}, \mathcal{C}_{u\to l}$ by $E_s(\mathcal{X}_{u\to l}), E_c(\mathcal{X}_{u\to l})$.
    \State Cycle reconstruct $\mathcal{X}_u$: $\mathcal{X}_{u \to l \to u} = G(\mathcal{C}_{u \to l}, \mathcal{S}_u)$.
    \State Let $A$ denote $E_s, E_c, G$.
    \State Let $B$ denote $D_{adv}^c, D_{adv}^x, D_{pre}^c, D_{pre}^x$.
    \State Fixing $A$, optimize $B$ by Eq.~\ref{eq11}.
    \State Fixing $B$, optimize $A$ by Eq.~\ref{eq10}.
\EndWhile
\State \Return $E_s, E_c$ and $G$.
\end{algorithmic}
\end{algorithm}

\textbf{Cycle Consistency Loss.} To guarantee that $\mathcal{X}_{u\to l}$ preserves the content of $\mathcal{X}_u$ while changing only the domain-specific style of label $L$, we apply a cycle consistency loss to the translator, which is defined as:
\begin{equation}
\begin{split}
\mathcal{L}_{cyc}^x =\mathbb{E}_{x_u\in\mathcal{X}_u}[\Arrowvert x_u - G(E_c(G(&E_c(x_u),s_r)),\\ &E_s(x_u))\Arrowvert _1].
\end{split}
\end{equation}

\textbf{Feature Consistency Loss.} To preserve the consistency of $\mathcal{C}_u$ and $\mathcal{S}_u$ in the representation level, we apply a feature consistency loss \cite{zhu2017toward} to the translator as follows:
\begin{equation}
\begin{split}
\mathcal{L}_{lat} &= \mathcal{L}_{rec}^c + \mathcal{L}_{rec}^s \\
 &= \mathbb{E}_{x_u\in\mathcal{X}_u}[\Arrowvert E_c(x_u) - E_c(G(E_c(x_u),s_r))\Arrowvert _1 \\
&+ \Arrowvert s_r - E_s(G(E_c(x_u),s_r))\Arrowvert _1]].
\end{split}
\end{equation}

\subsection{Optimization and Inference for AGUIT}
In the training phase, we optimize the style encoder $E_s$, content encoder $E_c$, generator $G$, predictors $D_{pre}^c, D_{pre}^x$ and discriminators $D_{adv}^c, D_{adv}^x$ for AGUIT jointly.
We can write the overall objective of $E_s, E_c$ and  $G$ as:
\begin{equation}
\begin{split}
\label{eq10}
\mathcal{L}_{G,E_c,E_s} &= 
\lambda_{cla}^s\mathcal{L}_{cla}^s
+ \lambda_{adv}^c \mathcal{L}_{adv}^c
- \lambda_{pre}^c \mathcal{L}_{pre}^c \\
&+ \lambda_{rec}^x \mathcal{L}_{rec}^x
+ \lambda_{adv}^x \mathcal{L}_{adv}^x
+ \lambda_{pre}^{x,G} \mathcal{L}_{pre}^{x,G} \\
&+ \lambda_{cyc}^x \mathcal{L}_{cyc}^x 
+ \lambda_{lat} \mathcal{L}_{lat}.
\end{split}
\end{equation}
The objective of $D_{adv}^c, D_{adv}^x$ and $D_{pre}^c, D_{pre}^x$ is defined as:
\begin{equation}
\begin{split}
\label{eq11}
\mathcal{L}_D = 
- \lambda_{adv}^c \mathcal{L}_{adv}^c
&+ \lambda_{pre}^c \mathcal{L}_{pre}^c \\
&- \lambda_{adv}^x \mathcal{L}_{adv}^x 
+ \lambda_{pre}^{x,D} \mathcal{L}_{pre}^{x,D},
\end{split}
\end{equation}
where $\lambda$ with different superscript and subscript are hyper-parameters for balancing the losses.
Alg.~\ref{alg1} summarizes the training process of AGUIT.
%

%
In the inference phase, a test image $x_t$ is encoded into $s_t$ and $c_t$ by $E_s$ and $E_c$.
Then, the attributes of $x_t$ can be changed by manipulating one or more dimensions of the style code $s_t$ optionally.
The $c_t$ and the manipulated style code $s_t^{'}$ are input into the generator $G$ and the manipulated style is translated while the content of original input $x_t$ is retained\footnote{More detailed description for the inference of AGUIT can be found in our supplementary materials.}.
%
\section{Experiments}
\label{sec:Experiments}
\subsection{Experimental Settings}
\textbf{Datasets.} We evaluate AGUIT on the Dog2Cat \cite{lee2018diverse} dataset which contains a variety of cat and dog faces, as well as the CelebA \cite{liu2015faceattributes} dataset which contains more than 200,000 labeled faces with attributes such as hair color, gender and presence of eyeglasses.
Among 40 attributes of CelebA dataset, we choose 8 most common attributes (\ie, Black Hair, Blond Hair, Brown Hair, Gender, Age, Smiling, Eyeglasses, Goatee) for experiments.
%

%
\textbf{Evaluation Metrics.} Firstly, to compare the quality of translated images, we conducted \textit{human preference} on Amazon Mechanical Turk (AMT) for evaluation.
The workers were given a source image with two target images translated by different models and required to answer the following questions, \ie, which target image has more precise attributes of the offered label, which target image retains more similar shape of source image and which target image looks more natural.
We randomly chose 100 source images.
For multi-modal translation, each source image has 19 random translations.
For multi-label translation, each source image has 30 label-specific translations.
Secondly, to measure the diversity, we compute the average \textit{LPIPS Distance} \cite{zhang2018unreasonable} between pairs of randomly sampled translation outputs with the same settings as \cite{zhu2017toward}.
\textbf{Baselines.} CycleGAN \cite{zhu2017unpaired} consists of two residual translation networks trained with adversarial loss and cycle reconstruction loss.
%
%
MUNIT \cite{huang2018multimodal} or DRIT \cite{lee2018diverse} decompose the image representation into content code and style code.
Then the translated image is generated by recombining the content code with a random style code sampled from the style space of target domain.
StarGAN \cite{Choi2017StarGAN} uses the images with labels as the input and manipulates the label to translate the corresponding attributes on the output images.
%
%
\begin{figure}[!t]
\centering 
\includegraphics[width=2.6in]{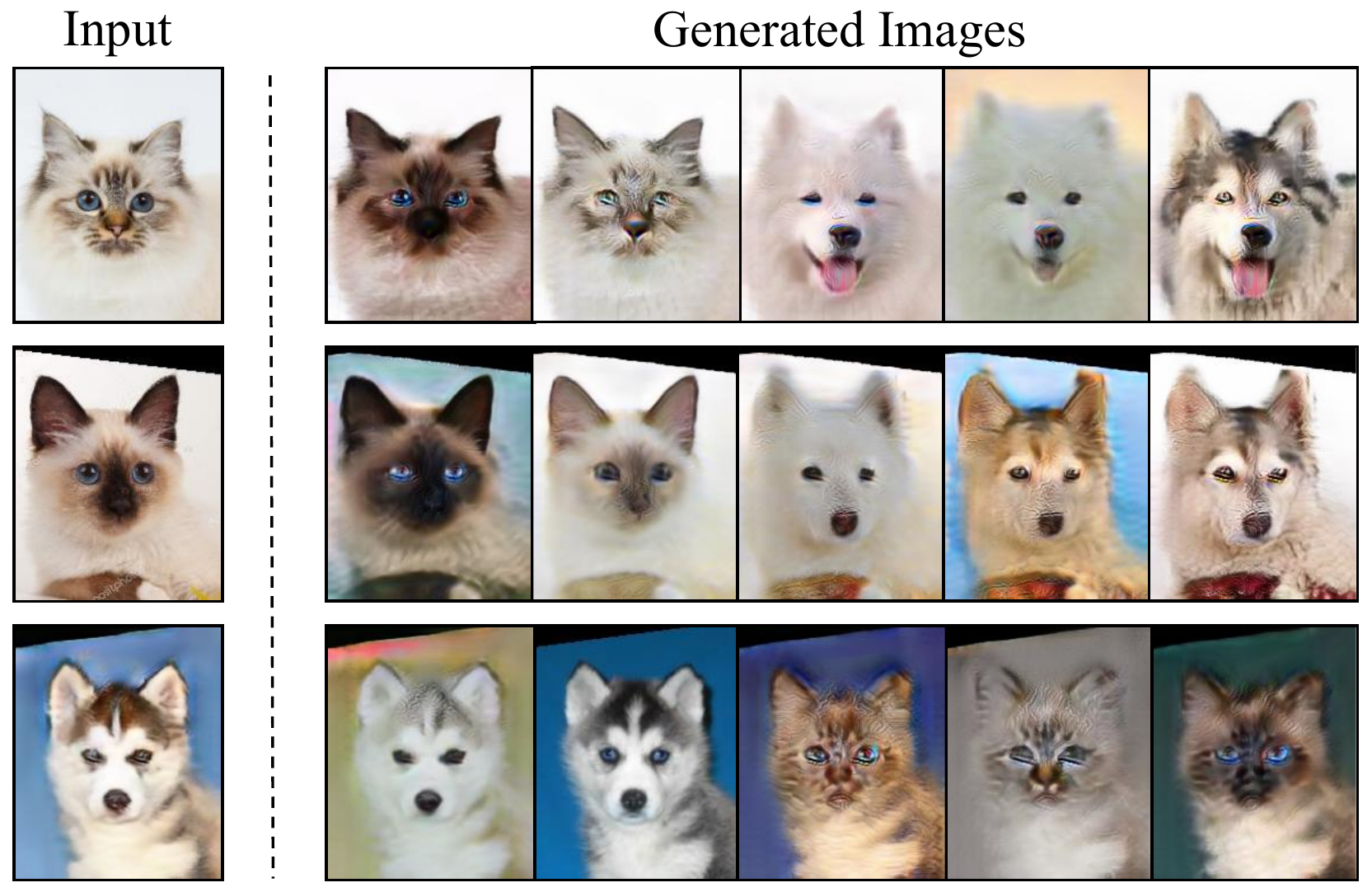}
\caption{
Examples of multi-modal translation with AGUIT on Dog2Cat dataset.
}
\label{fig.multi-modal}
\end{figure}
\begin{figure}[!t]
\centering %
\includegraphics[width=1\linewidth]{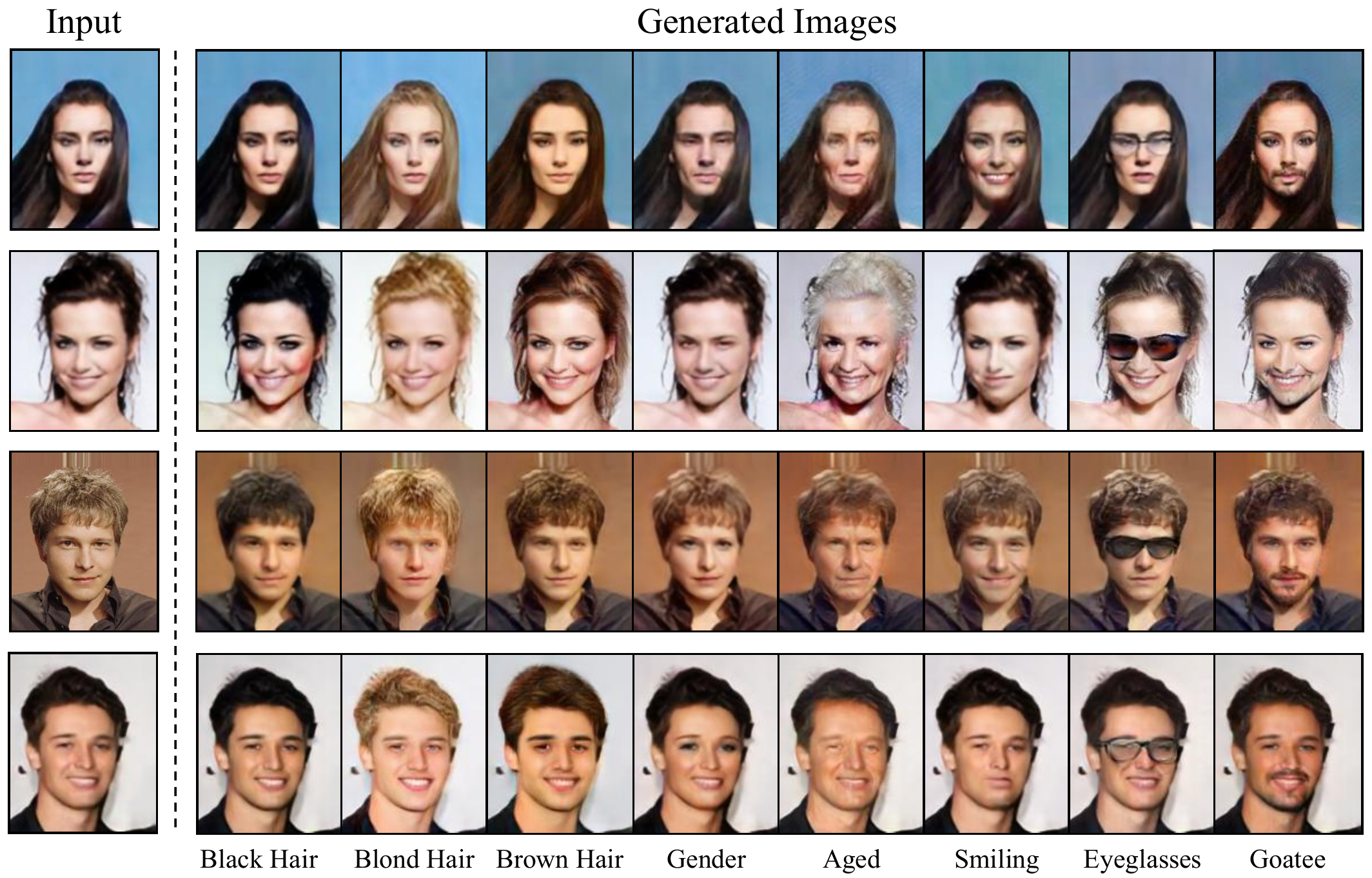}
\caption{
Examples of multi-domain translation with AGUIT on CelebA dataset.
}
\label{fig.multi-domain}
\end{figure}

\textbf{Implementation Details.} We use the input image size of $128\times128$ for all our experiments.
More results of $128\times128$ and $512\times512$ inputs, detailed settings of hyper-parameters, network architectures and optimizers can be found in our supplementary materials.
\subsection{Basic Image Translation Tasks}
We show that AGUIT can do basic image translation tasks (\ie, multi-modal translation and multi-domain translation) under the unpaired setting.

For multi-modal translation, the results are shown in Fig.~\ref{fig.multi-modal}, from which we can see that AGUIT produces a variety of results with the same or different species.
The translated images greatly retain the content (\ie, shape and posture) of input images.
For multi-domain translation, we change one of the labeled attributes, and the outputs are shown in Fig.~\ref{fig.multi-domain}.
From the results we can see that AGUIT learns attributes for the style code under the supervision of domain labels.
\begin{figure}[!t]
    \centering
    \subfigure[Multi-modal translation]{
    \label{Fig.ours-munit.a} 
    \includegraphics[width=3.0in]{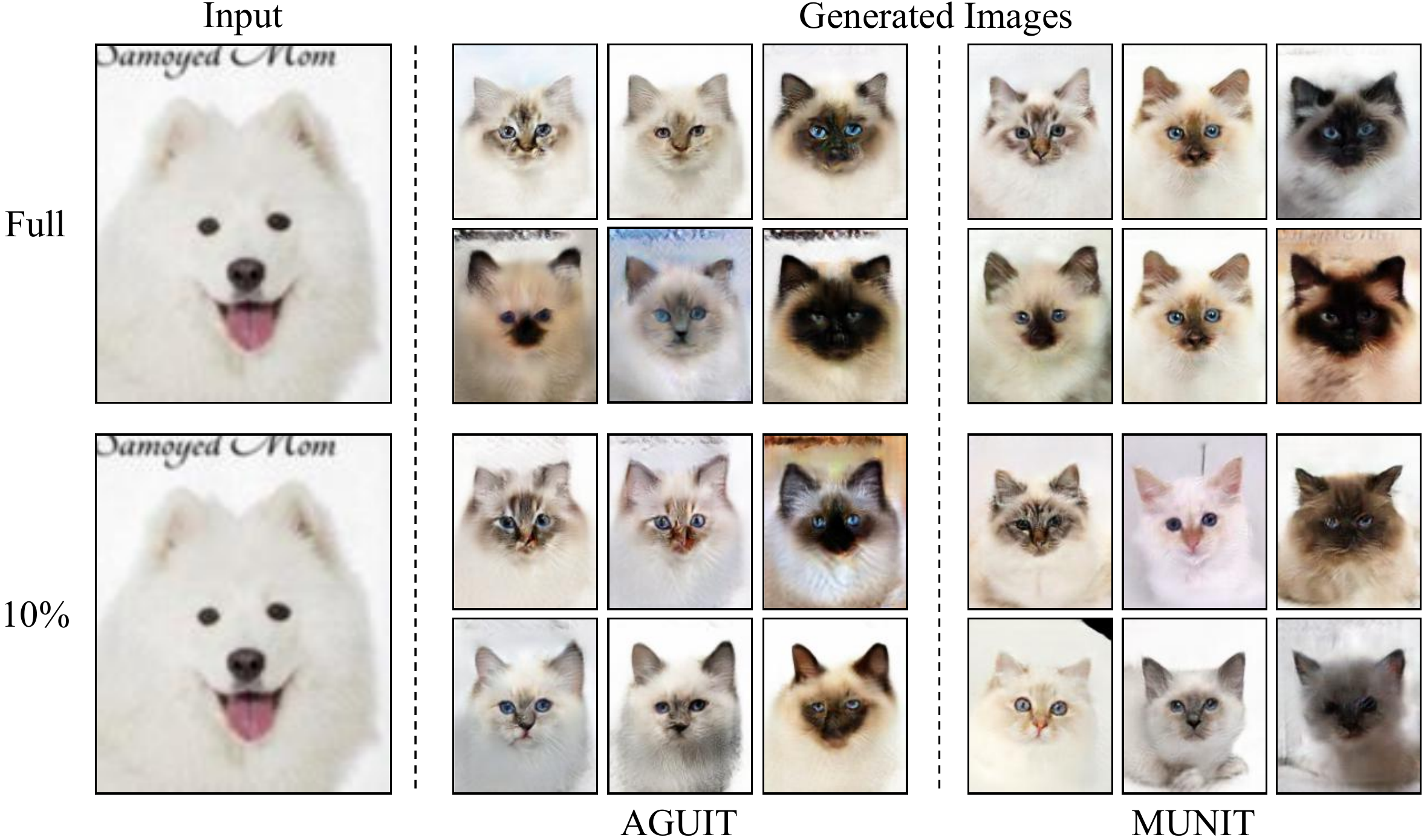}
    }
      \vspace{-0.1in}
    
    \subfigure[Multi-domain translation.]{
    \label{Fig.ours-munit.b}
    \includegraphics[width=3.0in]{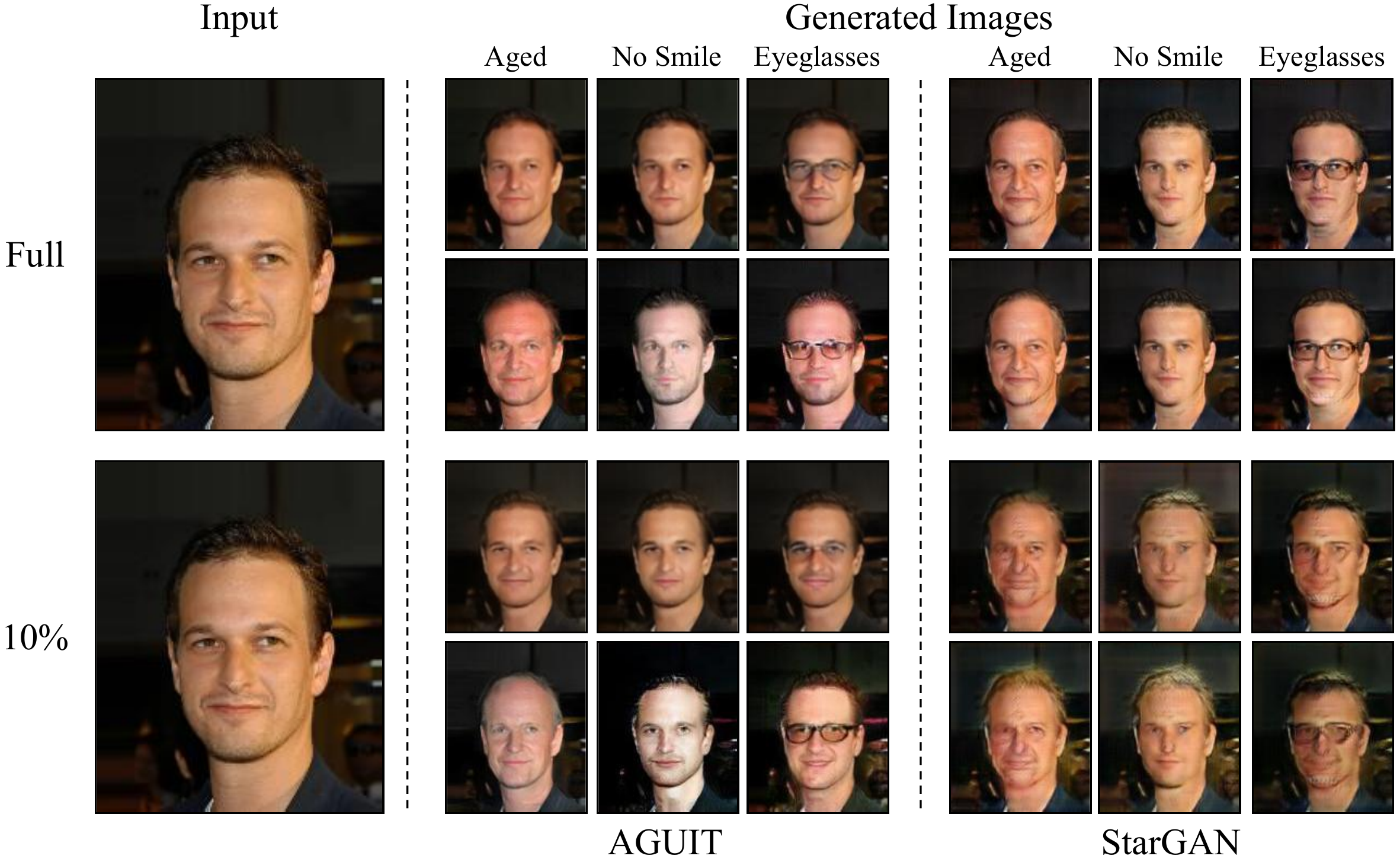}
    }
    \caption{Examples of basic translation tasks under semi-unsupervised setting.}
    \label{fig:ours-munit}
\end{figure}
\begin{table}[!t]
    \begin{center} 
        \subtable[Multi-modal translation]{ 
            \scalebox{0.58}[0.58]{
        \begin{tabular}{|c|c|c|}
            \hline
            Method-Settings & Quality  & Diversity \\ \hline
            MUNIT\cite{huang2018multimodal}-10\%OL & 9.3\% & \textbf{0.519} \\
            AGUIT-10\%OL & 13.3\% & 0.327 \\
            AGUIT-10\% & \textbf{27.0\%} & 0.442 \\
            \hline
            MUNIT\cite{huang2018multimodal}-full & 24.9\% & \textbf{0.492} \\
            AGUIT-full & \textbf{50.0\%} & 0.438 \\
            \hline
    \end{tabular}
    \label{Tab.sub.1}
    }
}
\subtable[Multi-domain translation]{
    \label{Tab.sub.2}
        \scalebox{0.58}[0.58]{
            \begin{tabular}{|c|c|c|}
                \hline
                Method-Settings & Quality & Diversity \\ \hline
                StarGAN\cite{Choi2017StarGAN}-10\%OL   & 20.2\% &  0.186 \\
                AGUIT-10\%OL & 23.8\% & 0.231\\
                AGUIT-10\% & \textbf{30.8\%} &  \textbf{0.252} \\
                \hline
                StarGAN\cite{Choi2017StarGAN}-full& 34.7\% &  0.165 \\
                AGUIT-full & \textbf{50.0\%} &  \textbf{0.242} \\
                \hline
        \end{tabular}}
    }
    \end{center}
    \caption{Quantitative evaluations of basic translation tasks under semi-supervised setting. The column of `Diversity' is the average LPIPS distance. The column of `Quality' is the human preference score. The setting of 10\%OL means only 10\% of training images are used. The setting of 10\% means 90\% training images are treated as unlabeled data for training.}
    \label{tab2}
\end{table}
\begin{figure}[!t]
\centering %
\includegraphics[width=1\linewidth]{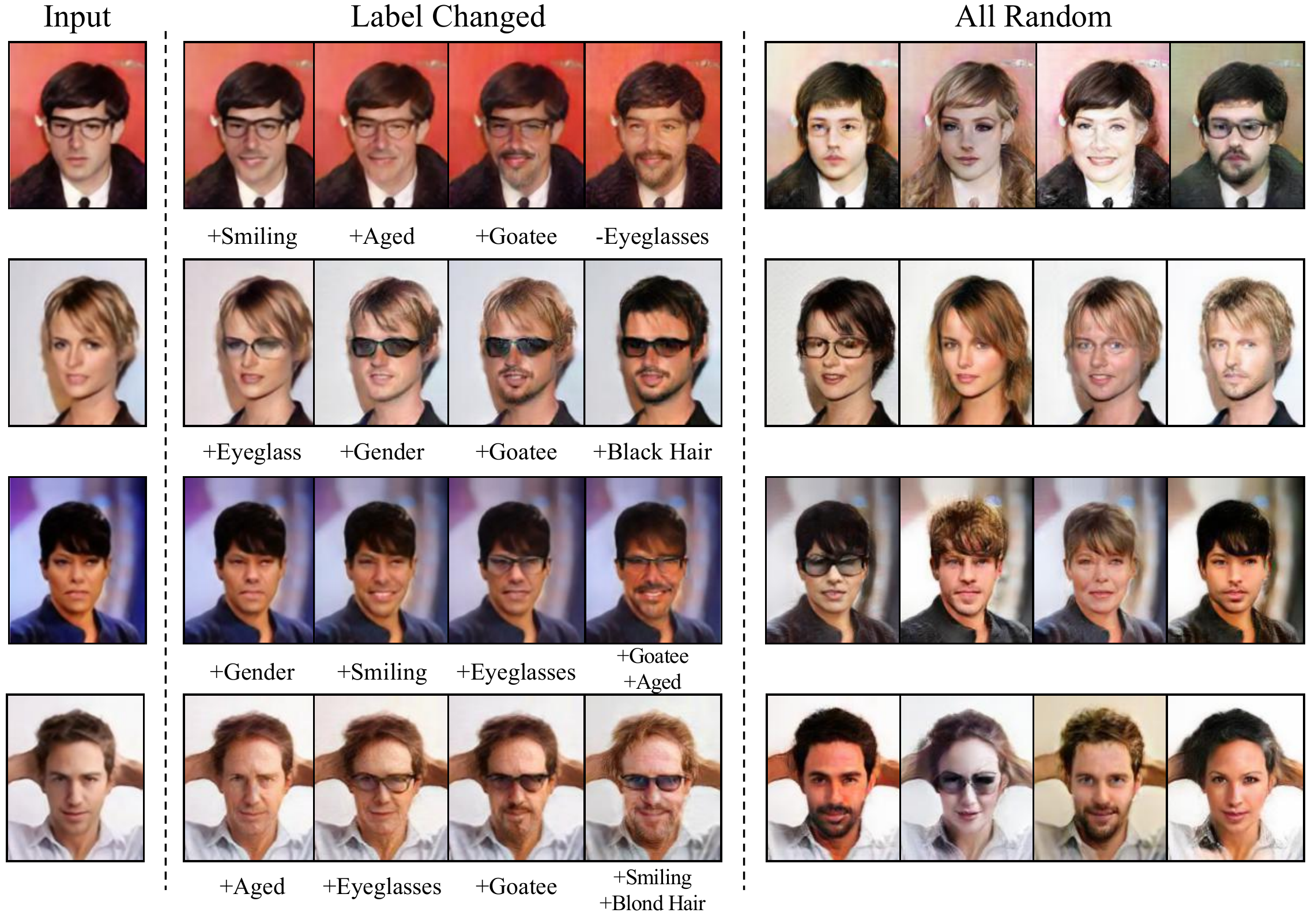}
\caption{
Examples of style specific translation.
The content codes of input images remain unchanged.
From the second column to the fifth column, we gradually add changes to the attribute part of style code.
From the sixth column to the ninth column, we randomly change all dimensions of the style code.
}
\label{fig.flexibility-1}
\end{figure}

\subsection{Benefits of Semi-supervised Setting}
In this section, we show the effectiveness of AGUIT under the semi-supervised setting, in which the training set contains labeled images mixed with unlabeled images.
We compare AGUIT with MUNIT and StarGAN under the settings of 10\% labels and full labels for multi-modal translation and multi-domain translation.
For AGUIT, we randomly choose images from labeled images to form the training pairs under the fully label setting.

\textbf{Qualitative Evaluation.} As shown in Fig.~\ref{Fig.ours-munit.a}, for multi-modal translation, AGUIT and MUNIT both have good results when using fully labeled images for training.
But when the number of labeled training images decreases drastically, the content of images generated by MUNIT change a lot compared to the input image.
On the contrary, AGUIT still generates a variety of results retaining the shape of input image.
As shown in Fig.~\ref{Fig.ours-munit.b}, for multi-domain translation, AGUIT and StarGAN both translate attributes well when using fully labeled images as the training set.
However, StarGAN cannot generate diversified results.
When the number of labeled training images decreases drastically, StarGAN fails both on the quality and diversity.
On the contrary, AGUIT accomplishes both goals well, as unlabeled images can be incorporated into the training phase and the data distribution is augmented.
\begin{figure*}[!t]
\centering
\includegraphics[width=6.2in]{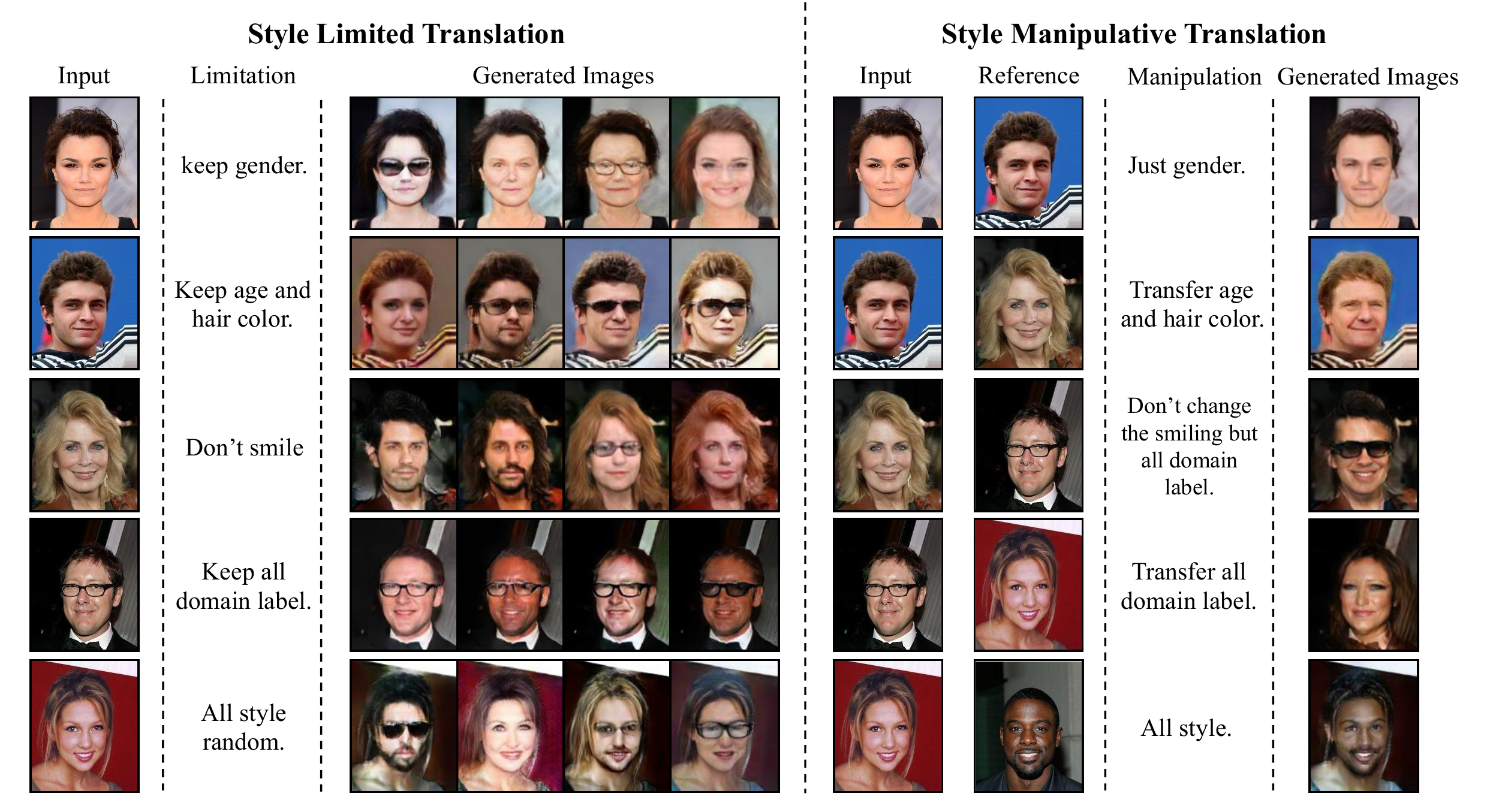}
\caption{
Examples of style limited translation (left) and style manipulative translation (right).
On the style limited translation, the content and style codes are given by the input images.
Some limitations are applied to the style code, and various output images are produced by AGUIT.
On the style manipulative translation, the content and style codes are given by the input images and style reference images, respectively.
Then manipulations are applied to the style code to translate the input images with the manipulated style.
}
\label{fig.flexibility}
\end{figure*}

\textbf{Quantitative Evaluation.} As shown in Tab.~\ref{tab2}\subref{Tab.sub.1}, for multi-modal translation, the human preference score of AGUIT is far beyond MUNIT, although MUNIT has a high value on LPIPS distance.
Commonly, high human preference score means high quality, and high LPIPS distance means high diversity.
We set high quality as prior to high diversity due to the goal of retaining the shape of input images.
From this perspective, AGUIT performs well compared with MUNIT.
As shown in Tab.~\ref{tab2}\subref{Tab.sub.2}, for multi-domain translation, AGUIT is superior to StarGAN on the evaluations of both quality and diversity.
For both tasks, the performance of AGUIT increases after incorporating unlabeled data.
To explain, the extra information of unlabeled data enriches the learning of representation.

Overall, based on the above qualitative and quantitative evaluations, AGUIT has stronger capability on the basic image translation tasks, and gains good benefits from the semi-supervised scheme.
\subsection{Style Operable Image Translation Tasks}
Because the style information is encoded into the style code that contains two variables drawn from the standard Gaussian distribution (\ie, the noise part) and the attribute’s distribution (\ie, the attribute part), AGUIT can accomplish high-level tasks and produce more plentiful results.
\textbf{Style Specific Translation.} As shown in Fig.~\ref{fig.flexibility-1}, we translate input image via style code, whose attribute part is assigned by changing specific dimensions.
The labeled attributes can be assigned cumulatively or randomly for translating input images.
The results show AGUIT translates images accurately with different specified style codes, while retaining the content of the input images.

\textbf{Style Limited Translation.} As shown in Fig.~\ref{fig.flexibility} left, we translate images via giving some limitations to the style code of input image.
From the results we can see AGUIT produces various results to meet the limitations.
In fact, an operation (\eg reverse, hold, random sample) can be applied to any dimensions of style codes unrelated with the limitations to enrich the output images.
%

\textbf{Style Manipulative Translation.} As shown in Fig.~\ref{fig.flexibility} right, we translate input images under style reference images with some manipulations.
The style reference image provides style code.
Then the manipulations can be assigned to it for translation.

Overall, the above three experiments show that AGUIT has the ability to handle the style operable image translation tasks and has great flexibility.
\subsection{Fine-controlled Image Translation Tasks}
In order to further reveal the ability of AGUIT, we conduct interpolation on both attribute part and noise part of style code for evaluating the effectiveness of disentanglement for AGUIT.
The results show that the representation learned by AGUIT is continuous and disentangled on both supervised attribute part and unsupervised noise part.
The outputs can be fine-controlled by adjusting the style code gradually.

\begin{figure}[!t]
\centering
\includegraphics[width=3.2in]{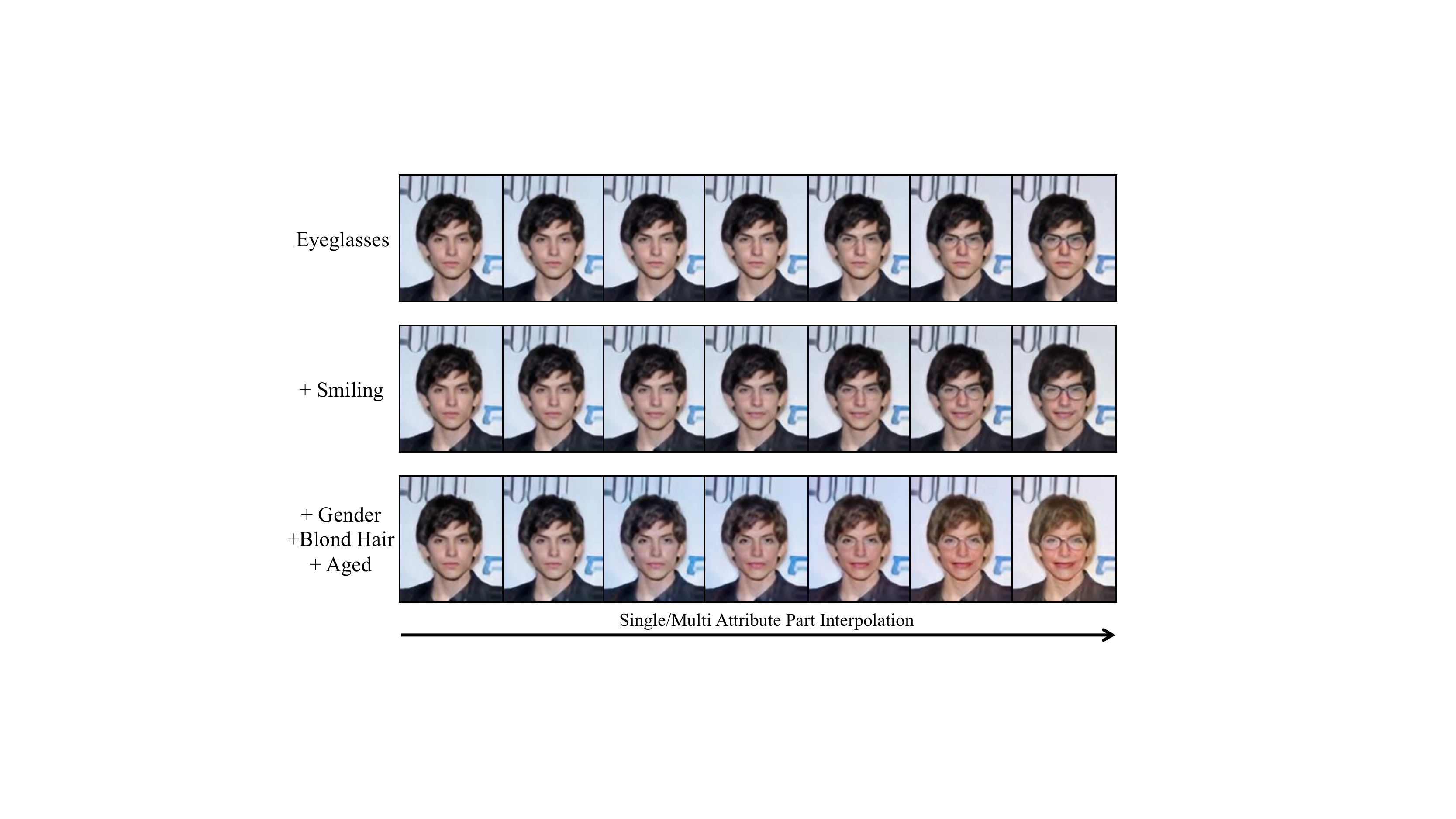}
\caption{
Supervised disentanglement on the attribute part of style code.
The interpolations are conducted from left to right.
Only one attribute is operated on the first row, and more attributes are added for the following rows.
}
\label{fig.intetpolatio-1}
\end{figure}

\begin{figure}[!t]
\centering
\includegraphics[width=3.2in]{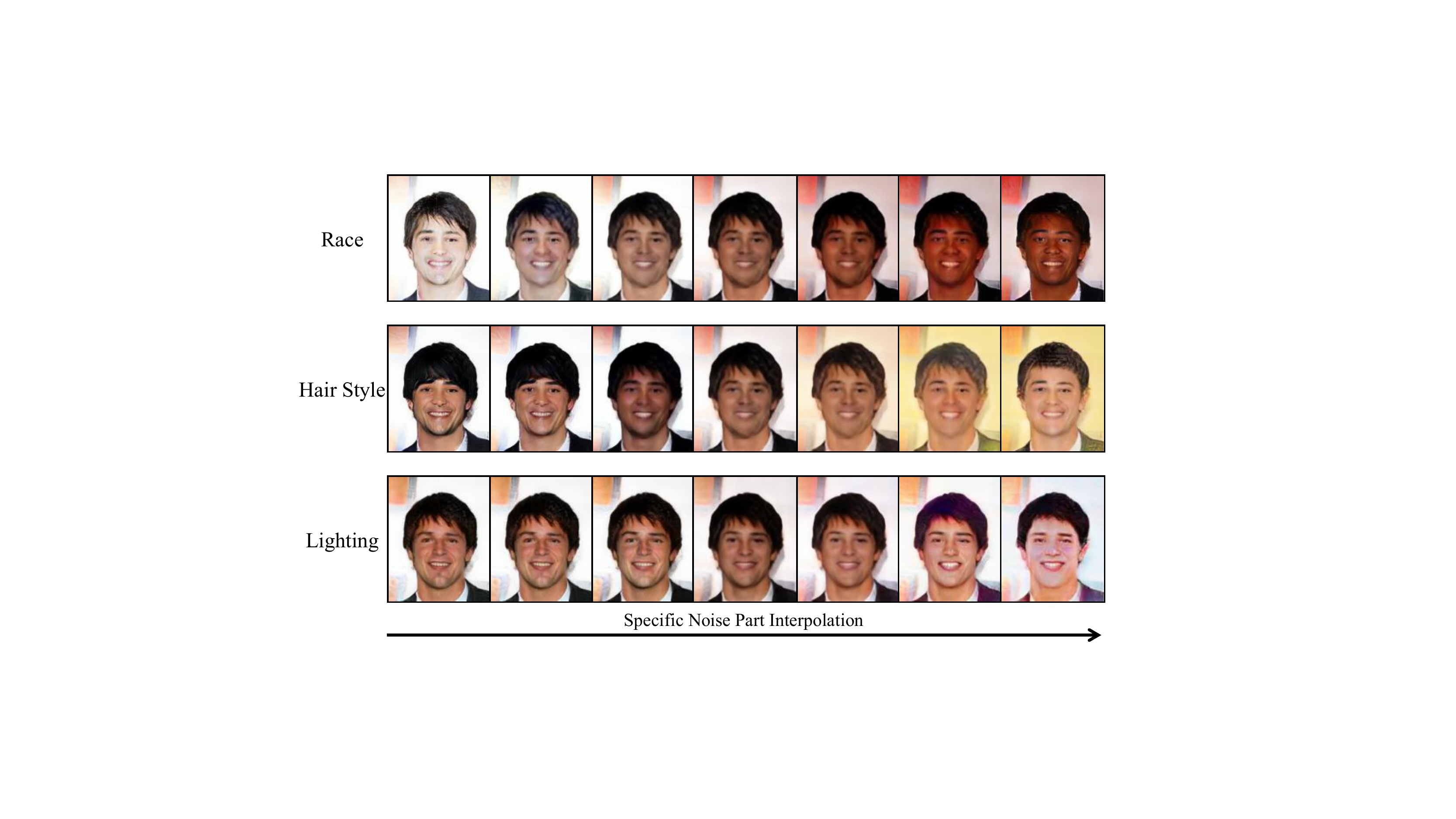}
\caption{
Unsupervised disentanglement on the noise part of style code.
The interpolations are conducted on single dimension from left to right on every row.
}
\label{fig.intetpolatio-2}
\end{figure}

\textbf{Supervised Disentanglement.} As shown in Fig.~\ref{fig.intetpolatio-1}, we can see that the labeled attributes are disentangled and can be fine-controlled while interpolating the value of the corresponding dimension.
The results suggest that whatever the single or multiple interpolation on the attribute part of style code, the representation space is independent and continuous.
The clean translated images indicate that the supervised disentanglement of style code has very high quality.
\textbf{Unsupervised Disentanglement.} Similarly, as shown in Fig.~\ref{fig.intetpolatio-2}, we can see that some unlabeled attributes such as lighting and hair style are learned and disentangled by AGUIT.
%
%
Though the background contains a little bit perturbation, the unsupervised disentanglement, to some extent, is a possible way to reduce the requirement of the attributes need to be labeled.

Overall, from the above experiments, AGUIT can achieve fine control for outputs because of the disentangled style code.
%

\subsection{Disentangled Transfer}
\begin{figure}[!t]
\centering
\includegraphics[width=3.0in]{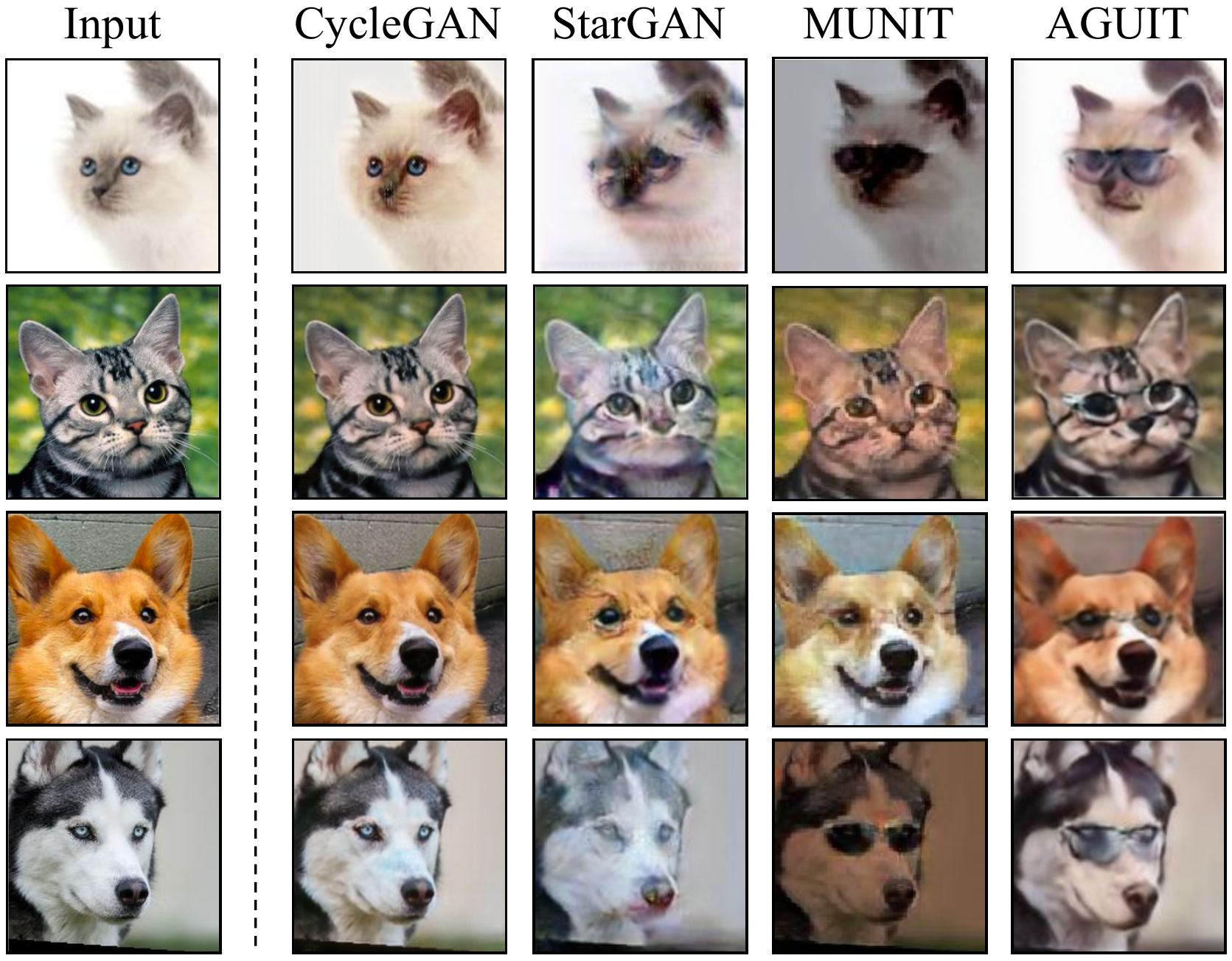}
\caption{
Example results of disentangled transfer.
}
\label{fig10}
\end{figure}

We further introduce a new task for image-to-timage translation, termed disentangled transfer.
Disentanglement is one of the goals of representation learning.
It is of great benefit for the practical application if the information learned by disentangled representation can be transferred.
For example, we can create zero-shot samples which has unrelated attributes.
To illustrate the goal of disentangled transfer clearly, we give an example as follows:
We utilize models trained on CelebA dataset to make the cats and dogs wear eyeglasses.
Because the eyes of dogs and cats are similar to human's, the attribute for human eyes ought to be transferred if the representation is disentangled well.
For comparison, we train CycleGAN and MUNIT by human faces with or without eyeglasses in CelebA dataset, and train the StarGAN by full CelebA dataset.
As shown in Fig.~\ref{fig10}, we can see that CycleGAN cannot put the eyeglasses to the face of dogs and cats.
We attribute it to the poor representation learning of CycleGAN.
StarGAN also fails on this task and brings noise to the translated images.
We attribute it to the fact that StarGAN is not capable for representation learning of attributes.
MUNIT has better results than the above models, where some translations achieve the goal while some are not, but the hue of the successful examples is changed drastically.
We attribute it to that the representation is not disentangled well, so that small turbulence on one dimension can change the whole image via the decoding scheme.
On the contrary, AGUIT works very well in this task.
Although it has a little changes on background, AGUIT finds the position of the eyes in the input image and puts the glasses on.
It suggests that AGUIT learns a good representation than other models, which also indicates the good potential of the disentangled transfer for image-to-image translation.

\section{Conclusion}
\label{sec:Conclusion}
In this work, we propose a model termed as AGUIT, which is the first model to consider UIT tasks with a semi-supervised setting, and achieves disentanglement of representation as well as fine control of outputs for UIT.
The semi-supervised learning scheme and the encoding of style information bring the capability to AGUIT.
Extensive experiments demonstrate the performance of AGUIT over the state-of-the-art image-to-image translation models.
First, we showed that AGUIT can do the basic image translation tasks (\ie, multi-modal and multi-domain translations).
Second, we qualitatively and quantitatively evaluated the benefits brought by the semi-supervised setting of AGUIT.
Third, we exhibited the style manipulation and fine control results.
Finally, we revealed that AGUIT can carry out disentangled transfer, which is a new task introduced to image-to-image translation.

{\small
\bibliographystyle{ieee}
\bibliography{egbib}
}

\onecolumn
\section*{A. Implement Details}

\begin{figure}[h!]
    \centering
    \includegraphics[width=0.82\linewidth]{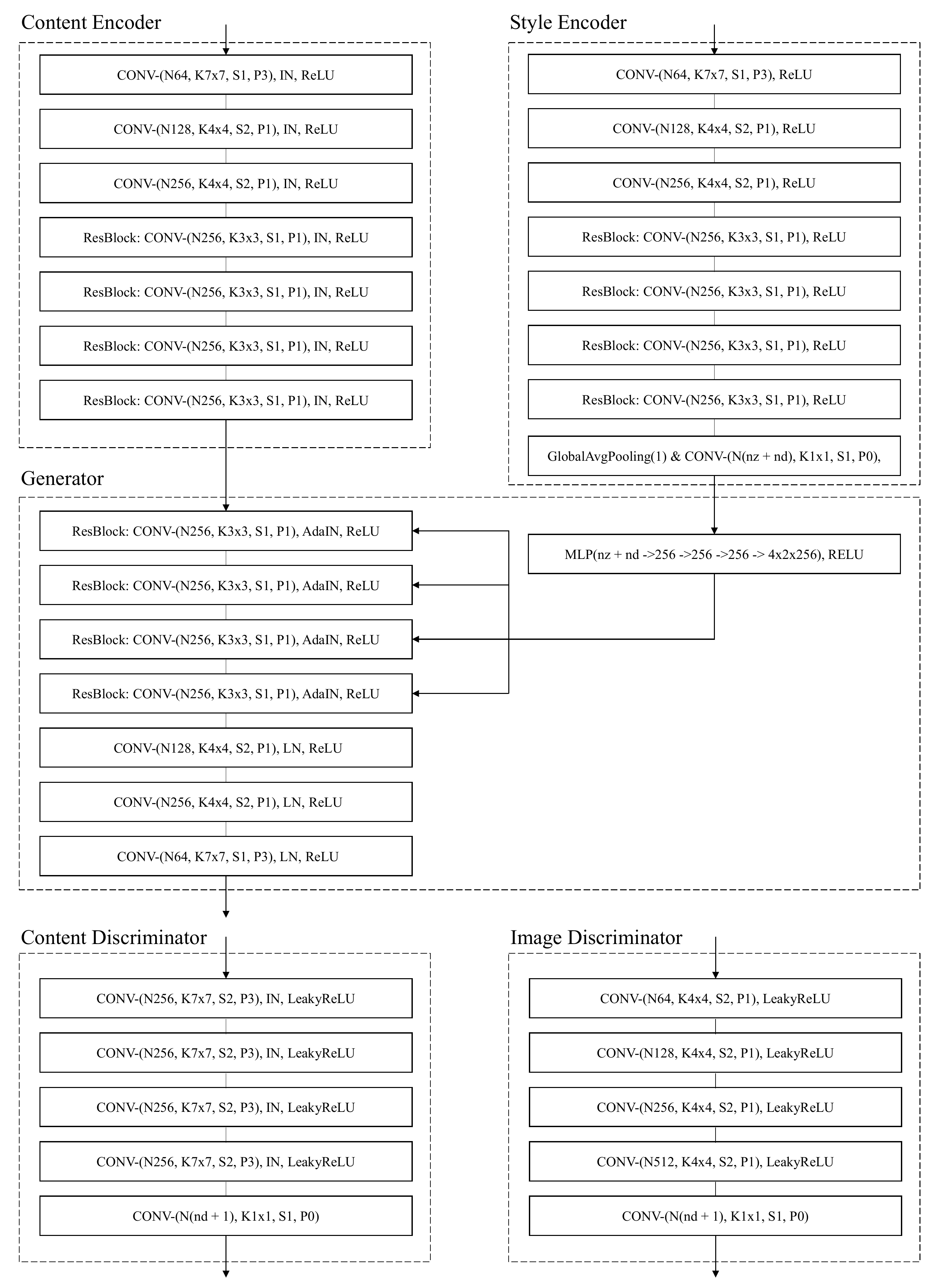}
    \caption{Detailed architecture of AGUIT. }
    \label{fig:architecture}
\end{figure}

\textbf{Architecture.}
The network architectures of AGUIT are shown as follows in Fig.~\ref{fig:architecture}. K: kernel size, S: stride size,
P: padding size, IN: instance normalization.
LN: layer normalization, AdaIN: adaptive instance normalization, 
nd: the dimension of attribute part, 
nz: the dimension of noise part.

\textbf{Training Details.}
For training, we use the Adam optimizer with a batch size of 8 for 128x128 photos of CelebA dataset and 1 for both Dog2Cat dataset and 512x512 photos of CelebA dataset. The learning rate is 0.0001, and exponential decay rates is set as $(\beta_1, \beta_2) = (0.5, 0.999)$. 
In all experiments, we set the hyper-parameters as follows:
$\lambda_{cla}^s = 10$, 
$\lambda_{adv}^c = 1$, 
$\lambda_{pre}^c = 1$, 
$\lambda_{rec}^x = 10$, 
$\lambda_{adv}^x = 1$, 
$\lambda_{pre}^{x} = 1$, 
$\lambda_{cyc}^x = 10$ and
$\lambda_{lat} = 10$, we use LSGAN for all GAN loss. 

\section*{B. Additional Experiment}

\begin{figure}[!h]
    \centering
    \includegraphics[width=0.82\linewidth]{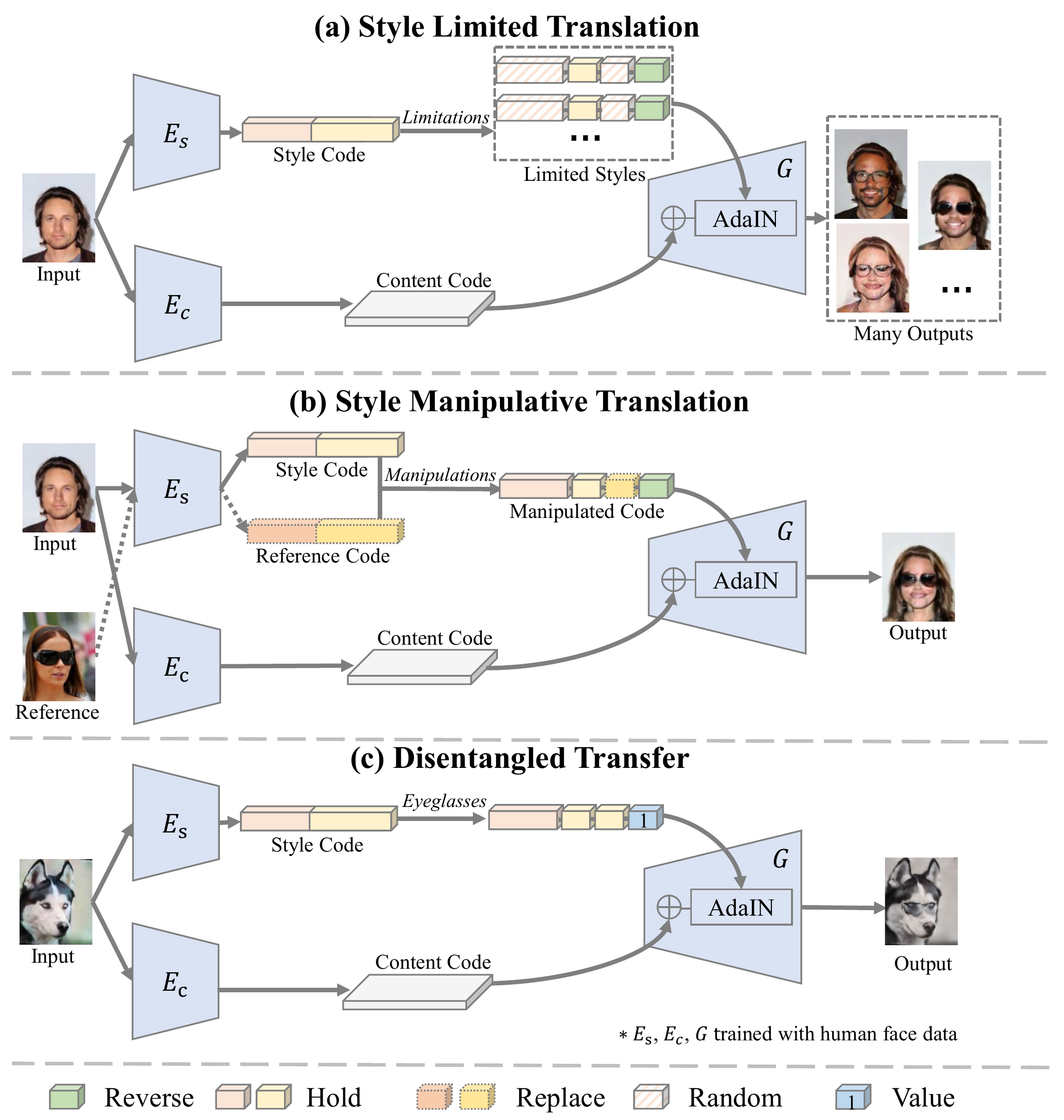}
    \caption{
    Detailed description of the inference phase of AGUIT.
    (a) Style code of input image is operated by Hold, Reversed or Random to get a variety of outputs. 
    (b) Style code of input image is operated by Hold, Reversed or replaced by
    reference's style code to get the manipulated output.
    (c) Style code of a dog is valued to +1 at the specific dimension for controlling the presence of eyeglasses. 
}
    \label{fig:diagram}
\end{figure}

The inference phase of AGUIT is shown in Fig.~\ref{fig:diagram}.
The operations for style code includes:
Hold: the same as original, 
Reverse: the opposite to original, 
Random: the random change, 
Replace: the replacement to a reference style code of another image,  
Value: the value to a specific number.

\begin{figure*}
    \centering
    \includegraphics[width=0.7\linewidth]{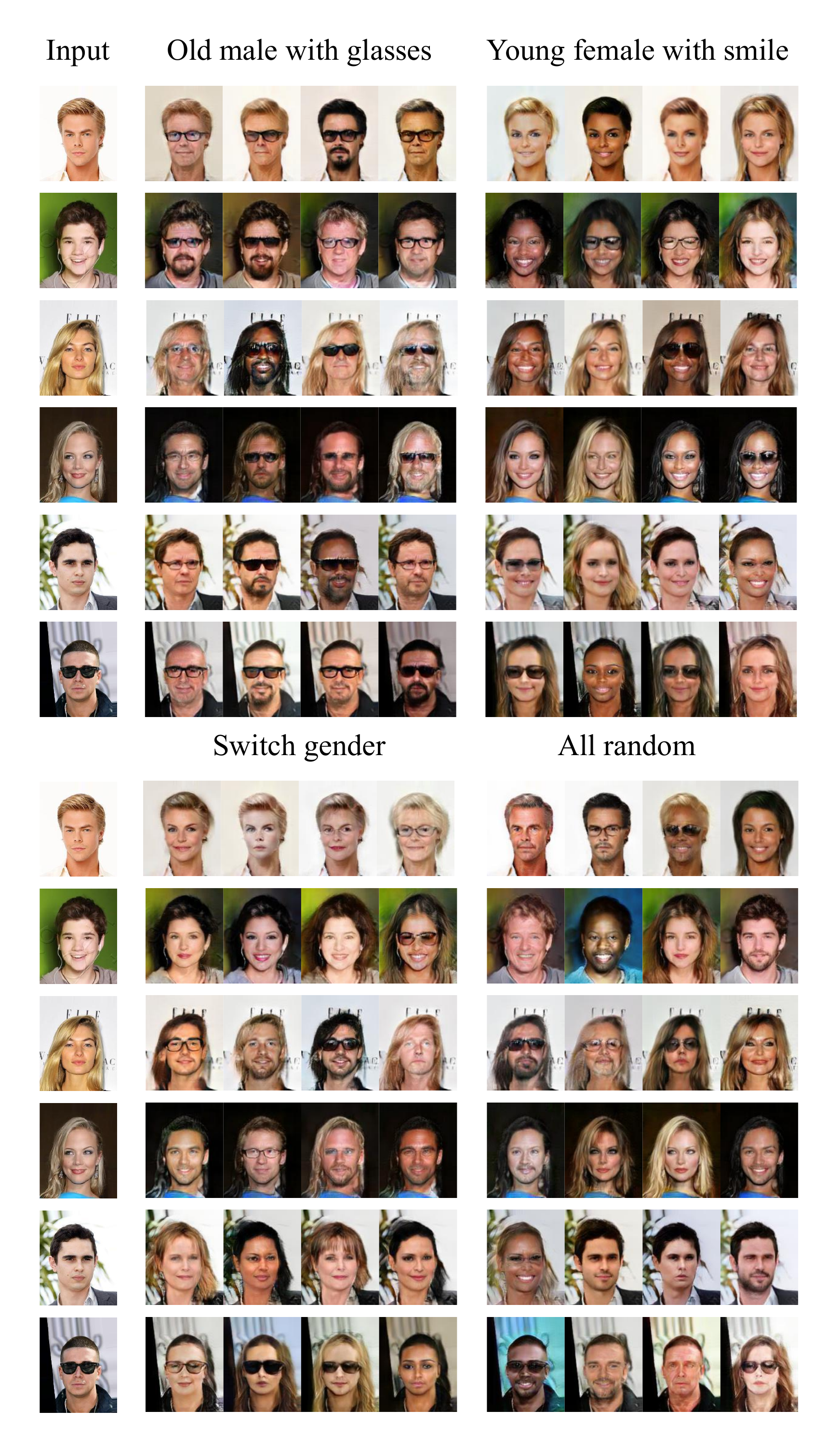}
    \caption{Additional results of AGUIT for style operable image translaiton tasks in CelebA dataset.}
    \label{fig:addition}
\end{figure*}

\begin{figure}[!t]
\centering
\includegraphics[width=1.1\linewidth]{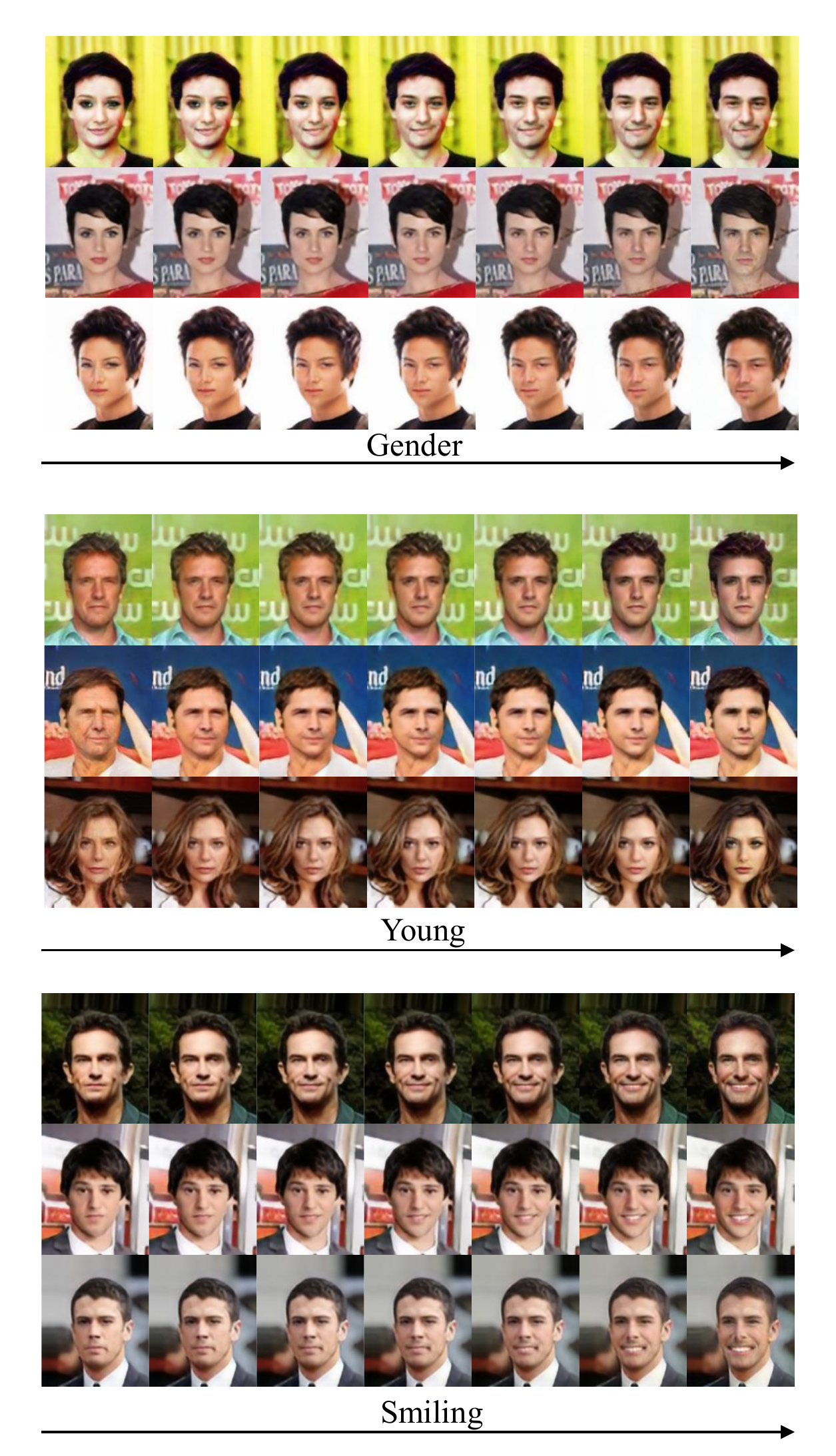}
    \caption{Supervised disentanglement on the attribute part of style code.}
\label{fig:fine1}
\end{figure}

\begin{figure}[!t]
\centering
\includegraphics[width=1.0805\linewidth]{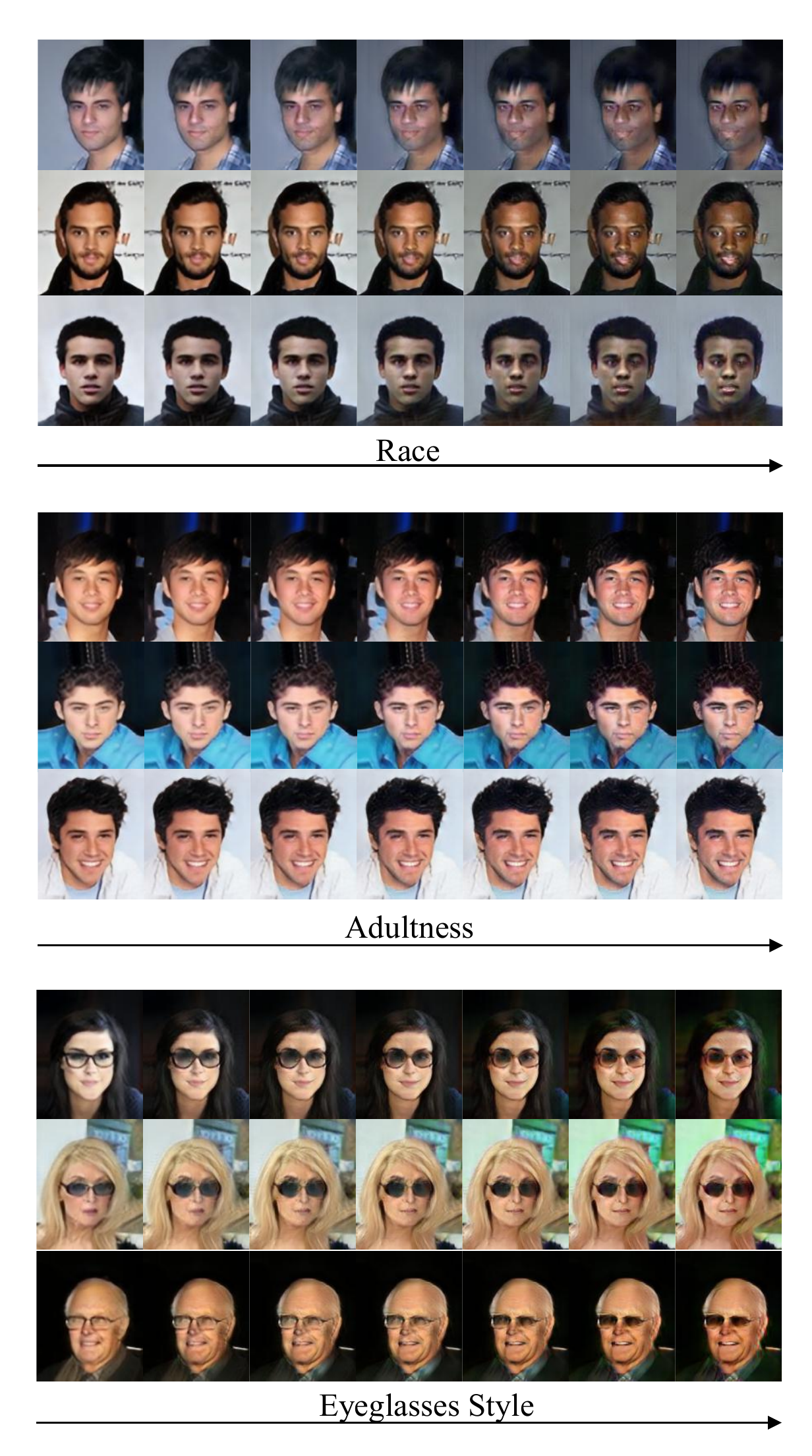}
    \caption{Unsupervised disentanglement on the noise part of style code.}
\label{fig:fine2}
\end{figure}

\begin{figure*}[!h]
\centering
        \includegraphics[width=1\linewidth]{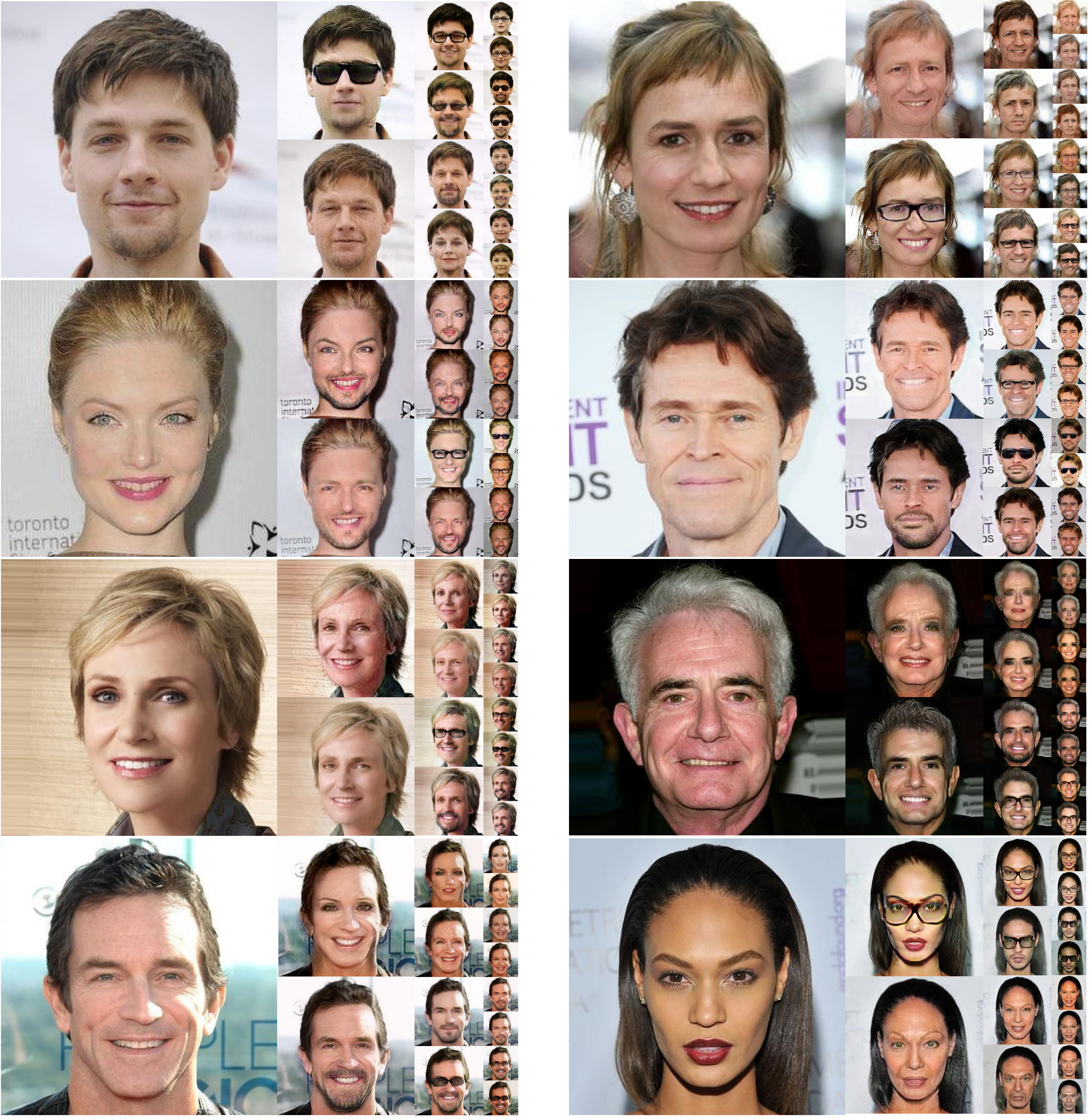}
    \caption{The results of 512x512 photos on CelebA dataset. First column is the input. More and more labels is reversed in the next two columns. The forth column is the multi-modal results.  Although limited batch size causes the unsupervised disentanglement of style code not as well as 128x128 models, the training of AGUIT does not need large-batch training like ProGAN, BigGAN, and StyleGAN.}
\label{fig:512}
\end{figure*}

In Fig.~\ref{fig:addition}, we show some additional results for style operable image translation tasks in CelebA dataset. In Fig.~\ref{fig:fine1} and Fig.~\ref{fig:fine2}, we show some additional results for fine-controlled image translation tasks with both noise part and attribute part of style code. 512x512 results are shown in Fig.~\ref{fig:512}.

\twocolumn

\onecolumn

\end{document}